\documentclass[journal]{IEEEtran}
\usepackage[left=0.625in,right=0.625in,top=0.75in,bottom=1in]{geometry}
\usepackage{ifpdf}
\usepackage{cite}
\usepackage{array}
\usepackage{dblfloatfix}
\usepackage{url}
\usepackage{amsmath}
\usepackage{ragged2e}
\usepackage{graphicx}
\usepackage{verbatim}
\usepackage{float}
\usepackage{lipsum}
\usepackage{multicol, blindtext}
\usepackage{caption}
\usepackage{subcaption}
\usepackage{xcolor}
\definecolor{mypink2}{RGB}{0, 0, 255}
\definecolor{green}{RGB}{0, 128, 0}
\usepackage{multirow}
\usepackage{array}
\usepackage{tabularx}
\usepackage[acronym]{glossaries}
\usepackage[ruled,vlined]{algorithm2e}
\usepackage{algorithmic}

\usepackage[utf8]{inputenc}
\usepackage{tcolorbox}
\usepackage{xcolor}
 \usepackage{tikz}
\usepackage{booktabs}   
\usepackage{multirow}   
\usepackage{siunitx}    
\usepackage{url}
\usepackage{hyperref}

\begin{document}

\title{\fontsize{12pt}{12pt}\selectfont LLM-Centric Agentic AI for UAV Swarms: Architecture, Enabling Technologies, and Open Problems}

\author{Yousef~Emami,~\IEEEmembership{Senior Member,~IEEE,}
        Rahim~Taheri,~\IEEEmembership{Senior Member,~IEEE,}
        Mohammadhossein~Homaei,~\IEEEmembership{Senior Member,~IEEE,}
        and~Muhammad~Atif Ur Rehman,~\IEEEmembership{Senior Member,~IEEE,}~and Mohammad~Shojafar,~\IEEEmembership{Senior Member,~IEEE}

\thanks{Copyright (c) 2026 IEEE. Personal use of this material is permitted. However, permission to use this material for any other purposes must be obtained from the IEEE by sending a request to pubs-permissions@ieee.org.}

\vspace{-20pt}}

\maketitle

\begin{abstract}
Uncrewed Aerial Vehicle (UAV) swarms have significant potential for applications such as Search and Rescue (SAR) and environmental monitoring, but their real-world deployment is limited by a lack of situational awareness, intermittent connectivity, and significant cybersecurity risks. Agentic Artificial Intelligence (AI) represents a shift from standalone Large Language Model (LLM) toward closed-loop cognitive architectures that integrate perception, memory, reasoning/planning, and action to enable adaptive, goal-directed swarm behavior. Within this framework, Agentic AI provides a unifying structure for autonomous and adaptive swarm operations while expanding the system’s attack surface compared to conventional AI systems. This paper proposes  LLM-Centric Agentic AI for UAV Swarms (LAUS) and reviews key enabling technologies such as onboard and edge 
computing, 5G/6G connectivity, multimodal intelligence, and cybersecurity mechanisms, and
analyzes threats such as Priority
Manipulation Attacks (PMA) that
can distort decision-making and degrade network performance. Finally, it identifies open research challenges, including hallucination-resistant reasoning, onboard LLM deployment under SWaP constraints, and standardized security benchmarks for 
perception-reasoning attacks in agentic UAV systems.

\end{abstract}

\IEEEpeerreviewmaketitle

\section{Introduction}

Thanks to their high mobility and ability to establish Line-of-Sight (LoS) communications, Uncrewed Aerial Vehicle (UAV) swarms have been widely adopted in Search and Rescue (SAR) missions, environmental monitoring, parcel delivery, precision agriculture, and construction \cite{kurunathan2023machine}. UAV swarms face several key challenges and limitations, including ethical concerns, poor energy efficiency that restricts operational endurance, difficulties in autonomous control and task allocation within complex environments, communication breakdowns as swarm size grows, insufficient robustness and scalability of algorithms for dynamic settings, synchronization issues caused by environmental disturbances, and security vulnerabilities that can lead to coordination failures\cite{10430396},\cite{10516683}.

\begin{figure*} [t]
    \centering 
    \captionsetup{justification=justified}
    \includegraphics[width=16cm, height=10cm]{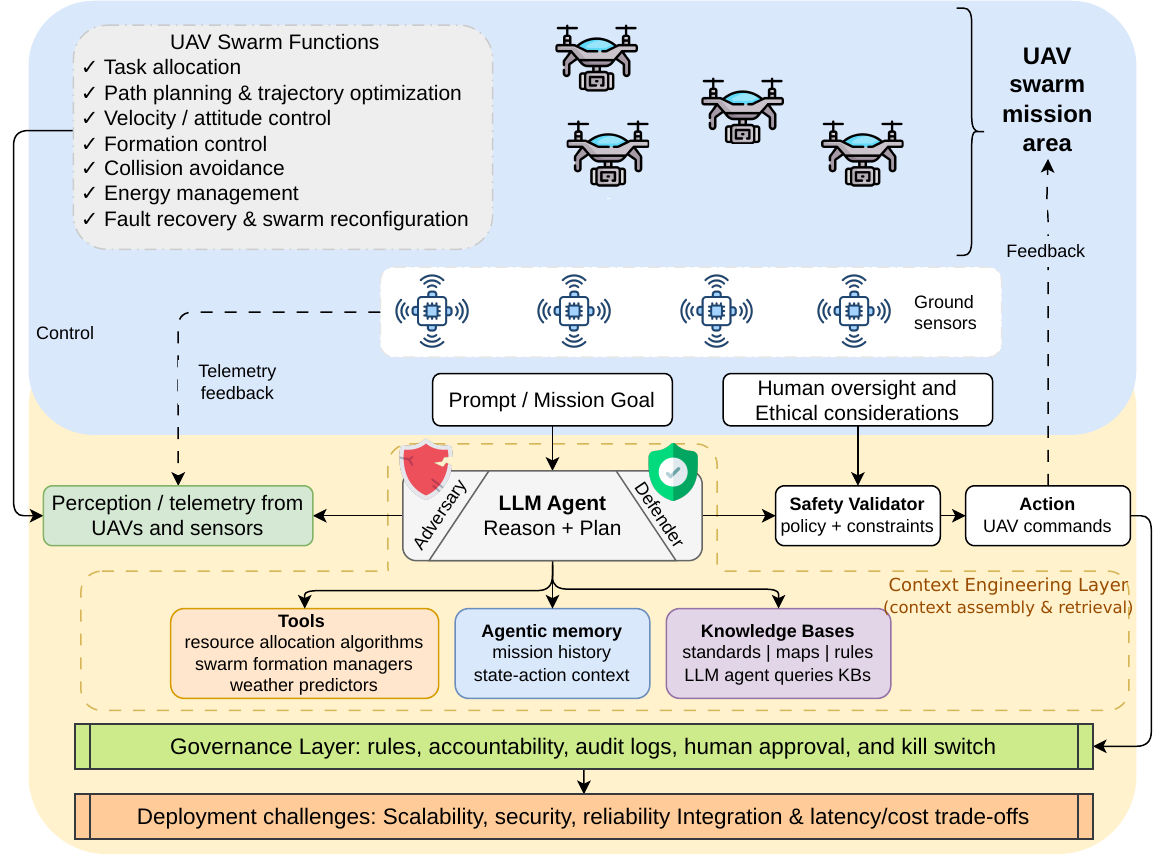}
    \caption{The proposed LAUS architecture, in which a context engineering layer assembles perception, memory, and knowledge-base inputs for the LLM Agent to translate mission objectives into actionable UAV commands via a closed-loop control cycle, operating under system-level deployment constraints (scalability, security, reliability, integration, and latency/cost).}
    \label{fig:digi}
\end{figure*}

Real-world UAV swarm deployment is fundamentally constrained by restricted situational awareness, intermittent connectivity, and persistent cybersecurity threats, challenges that standard LLMs cannot adequately address. Despite their strong language understanding, LLMs are architecturally misaligned with autonomous swarm operations, lacking temporal memory, native tool interaction, and the capacity to generate verifiable real-time control actions in safety-critical settings. Agentic AI addresses this gap by shifting from passive inference to closed-loop autonomy, in which an Agent Loop continuously perceives, plans, executes, and learns, enabling adaptive and self-improving swarm behavior. Crucially, the focus of Agentic AI is the cognitive architecture within each agent, not merely the coexistence of multiple cooperating UAVs\cite{11370176}.
\par
Sapkota \textit{et al.} \cite{sapkota2025} provide a foundational framework that defines and differentiates Agentic UAVs from traditional ones, while mapping their transformative potential across seven key societal domains. Koubaa \textit{et al.} \cite{koubaa2025} suggest a five-layer Agentic UAV framework structured around perception, reasoning, action, integration, and learning. This paper proposes LLM-Centric Agentic AI for UAV Swarms (LAUS), a closed-loop LLM-assisted framework that translates mission objectives into coordinated UAV actions. It reviews enabling technologies such as edge computing, 5G, Multimodal LLMs (MLLMs), and cybersecurity, and analyzes key threats, including prompt injection, memory poisoning, and adversarial manipulation. A case study on Priority Manipulation Attacks (PMA) demonstrates how corrupting input observations can mislead agent decisions and degrade swarm performance, highlighting the need to secure the perception–reasoning interface.

The key contributions are summarized as follows:

\begin{itemize}

\item We propose LAUS, a novel closed-loop agentic framework where an LLM Agent handles swarm functions by translating high-level mission objectives into actionable UAV commands. We also review its critical deployment enablers, onboard and edge computing, 5G/6G connectivity, MLLMs, 
and cybersecurity frameworks. 

\item We investigate emerging adversarial threats against LAUS, including prompt injection, memory poisoning, retrieval poisoning, malicious tool feedback, and adversarial agent behavior, and discuss defense-in-depth mechanisms spanning perception, memory, tool use, reasoning, and action execution.

\item We present a novel case study of a PMA against an LLM-assisted UAV data collection scheduling agent, demonstrating how adversarial manipulation of observations can influence agent decisions and degrade network performance, thereby exposing a new attack surface at the perception--reasoning interface.

\end{itemize}

The remainder of this paper is organized as follows. Section \ref{sec2} presents the proposed LAUS, along with its deployment challenges and applications. Section \ref{sec3} presents enabling technologies, onboard/edge computing, 5G communications, MLLMs, and cybersecurity. Section \ref{sec4} discusses adversarial threats and defense mechanisms for LAUS. Section \ref{sec5} presents a case study on PMA. Section \ref{sec6} discusses challenges and open problems. Finally, Section \ref{sec7} concludes the paper.

\section{Proposed LAUS} \label{sec2}
This section presents the proposed LAUS architecture, a closed-loop framework that integrates LLM-assisted reasoning with context engineering, tool calling, agentic memory, governance, and human oversight to enable trustworthy UAV swarm operations. The architecture transforms high-level mission goals into coordinated swarm behaviors through continuous perception, reasoning, planning, action, and learning while leveraging external tools for sensing, knowledge retrieval, resource optimization, and control. To support long-duration and dynamic missions, LAUS incorporates agentic memory for adaptive knowledge management and a governance layer that enforces safety, accountability, transparency, and Human-in-the-Loop (HITL) decision-making. The section further discusses the deployment challenges and applications of agentic UAV swarms and highlights the broad applicability of the proposed framework across diverse real-world UAV applications.
\par
Fig. \ref{fig:digi} depicts the proposed LAUS, where the process begins with collecting and processing raw sensory data and shaping a high-level prompt fed into the LLM Agent, which engages in explicit reasoning and planning to translate abstract objectives into actionable strategies. Within this framework,
prompt engineering focuses only on single-turn instructions while Context Engineering (CE) is a system-level discipline that determines what information an LLM sees, remembers, and uses across multiple steps. CE acts as the architectural layer that turns stateless LLMs into agentic systems by managing memory, retrieval, context assembly, and tool use. It helps maintain coherence in long-running tasks like data collection in UAV swarms and reduces issues like context overload, cost, and information loss.  This system-level role of CE is further operationalized through tool calling, which is a key enabler of LAUS, completing complex tasks by dynamically accessing and acting upon external resources.
\par
Tool calling serves as a key execution mechanism within LAUS, operationalizing every stage of the closed-loop architecture. In perception, tools interface with onboard multimodal sensors (LiDAR, thermal/multispectral cameras, GNSS/IMU) and edge analytics pipelines to acquire and parse raw telemetry into structured environmental observations, such as channel quality, residual energy of ground sensors, and inter-UAV proximity, forming the situational awareness that grounds all subsequent swarm decisions. In reasoning/planning, the agent invokes specialized retrievers, querying knowledge bases for 3GPP channel standards, and leveraging dynamic knowledge graphs to decompose high-level mission directives into distributed, multi-UAV sub-tasks with collision-free coordination. In action, tool calling translates reasoning outputs into deterministic cyber-physical commands, invoking resource allocation algorithms for fair sensor data collection based on queue backlog and energy budgets, swarm formation managers for dynamic topology reconfiguration, weather predictors for gust-aware trajectory adaptation, and differential GNSS engines for precise localization in GPS-challenged environments, with all outputs gated through independent safety validators before reaching the flight controller to prevent unsafe actuation. Beyond perception, reasoning, and action, maintaining continuity across missions requires a persistent memory mechanism that allows agents to retain and reuse prior experiences.
\par
In memory, tools provide the persistent read/write 
interface to external modules, such as distributed 
vector databases and edge caches, enabling agents to 
retrieve historical mission telemetry. This 
closed-loop tool-calling regime ensures that LAUS 
agents not only react to immediate conditions but also 
continuously evolve their swarm-level intelligence, 
sustaining episodic coherence and adaptive 
self-improvement even under intermittent connectivity 
and adversarial perturbations\cite{koubaa2025}. 
However, the rigidity of current memory architectures 
undermines this adaptability: existing systems provide 
only basic storage and retrieval, relying on manual 
predefined structures, fixed workflows, and rigid 
retrieval rules. Even graph-based approaches that 
improve organization using Retrieval-Augmented 
Generation (RAG) and structured databases still depend 
on predefined schemas, making it difficult for agents 
to reorganize knowledge dynamically or generalize 
effectively in open-ended environments. This limitation has motivated the development of more flexible, agentic memory paradigms that move beyond static storage and retrieval mechanisms.
\par
To address this, emerging agentic memory frameworks 
represent information as atomic, interconnected notes 
with structured attributes and embedding vectors for 
semantic similarity matching, enabling autonomous 
knowledge reorganization without static predefined 
schemas\cite{xu2026mem}. In practice, such frameworks 
immediately encode new experiences as structured notes, dynamically link them to historically similar 
situations, and continuously update existing knowledge 
through emergent semantic associations, enabling agents to maintain autonomous decision-making during 
communication blackouts by relying on locally cached, 
historically grounded insights and inferred peer-state 
models. While these advances strengthen the cognitive capabilities of LAUS, safe real-world deployment further requires explicit governance mechanisms to regulate autonomous behavior.
\par
Only validated commands are translated into precise UAV commands that drive the underlying UAV swarm functions, which encompass a comprehensive suite of low-level capabilities as depicted in Fig. \ref{fig:digi}. Throughout this entire pipeline, human oversight and ethical considerations are deeply embedded, ensuring that autonomy does not compromise moral or legal accountability. This is reinforced by a dedicated Governance Layer that improves system trustworthiness through enforceable rules, comprehensive audit logs for end-to-end traceability, mandatory human approval workflows for critical decisions, and a fail-safe kill switch that provides ultimate emergency override capability, ensuring that human agency remains the final authority over autonomous swarm operations. Within this governance structure, HITL oversight plays a critical role in ensuring that high-stakes decisions remain aligned with ethical and operational requirements. Effective governance is crucial for managing large autonomous UAV swarms, as it ensures a balance between operational autonomy and strict oversight. Organizations must establish clear KPIs such as communication reliability, collision avoidance, energy efficiency, and mission success rates to evaluate performance. Because these systems may operate in sensitive or restricted environments, strong risk-based controls are required, including continuous monitoring, anomaly detection, audit logging, secure communications, and emergency kill-switch mechanisms. In addition to this governance layer, HITL mechanisms ensure continuous human supervision during UAV swarm operations.

\par
HITL mechanisms remain essential in UAV swarm operations. Human operators provide strategic oversight, ethical judgment, and mission-level decision-making that autonomous agents cannot reliably achieve. In scenarios involving civilian safety, conflict zones, or uncertain operational conditions, human intervention can prevent harmful or unintended actions caused by hallucinations, goal drift, or incorrect reasoning by AI agents. Human supervision also improves public trust by ensuring that responsibility for critical decisions remains with accountable operators and organizations, rather than with the autonomous system. Beyond individual decision oversight, large-scale swarm deployment also demands system-level governance to manage performance, risk, and accountability across missions.
\par
Ethical considerations further reinforce the need for transparency and explainability in LAUS. Autonomous swarm behavior can become difficult to interpret when multiple agents independently adapt their actions based on local observations and collaborative reasoning. This creates challenges in understanding why specific decisions were made, particularly during mission failures or unexpected events. Incorporating causal reasoning, explainable AI mechanisms, and transparent system architectures can improve observability and allow operators to trace swarm decisions more effectively. These considerations collectively motivate the deployment challenges examined next. Despite these advances in control, memory, and interpretability, real-world deployment remains constrained by broader system-level limitations.
\par
The deployment of agentic AI in UAV swarms is limited less by model capability and more by system-level constraints such as scalability, security, reliability, integration complexity, and latency/cost trade-offs. Addressing these challenges requires end-to-end engineering with strong governance, monitoring, and safety controls. Nevertheless, LAUS demonstrates strong potential across a wide range of real-world domains. Across fields such as agriculture, disaster response, infrastructure inspection, logistics, and environmental monitoring, its value lies in enabling scalable, intelligent coordination that improves efficiency, situational awareness, and operational precision\cite{sapkota2025}.

\par
Having established the conceptual architecture of Agentic AI, its closed-loop control cycle, tool calling features, and deployment challenges along with applications, the natural question becomes: what underlying technical infrastructure makes such systems feasible on resource‑constrained UAV platforms? The following section details each of these enabling technologies.

\section{Enabling Technologies} \label{sec3}
To operationalize LAUS, the conceptual agent loop must be supported by a layered technical infrastructure. The following enabling technologies, onboard computing, edge-based deployment, 5G/6G connectivity, MLLMs, and cybersecurity collectively bridge the gap between abstract reasoning and tangible swarm operation.

\subsection{Onboard Computing}

Onboard computing is the most constrained scenario, where LLM agents operate entirely on a single UAV. This approach removes reliance on external communication links and improves data privacy, but it is severely limited by the storage, compute, and energy constraints typical of small aerial platforms. Recent LLMs require prohibitive resources: even moderately sized models with billions of parameters need high-end GPU clusters for fine-tuning and inference. Although compression techniques such as quantization, pruning, and knowledge distillation can reduce onboard resource requirements, they often degrade the reasoning and contextual understanding necessary for autonomous navigation, surveillance, and decision-making. As a result, onboard computing may lead to lower mission success rates and reduced swarm-level performance, making it impractical for real-time, mission-critical UAV operations.

\subsection{Edge-Based Deployment}

Edge-based deployment has emerged as a compelling middle ground, bringing LLM agents closer to data sources, such as UAVs or nearby edge infrastructure. This approach leverages model optimization techniques, including quantization, pruning, and parameter-efficient fine-tuning, to adapt LLMs for resource-constrained onboard hardware, enabling real-time inference with near-instantaneous responsiveness. Recent advances in distributed edge computing frameworks further improve feasibility by using batching and adaptive quantization to achieve high-throughput inference on commercial-grade devices.

\subsection{5G/6G and Next-Generation Connectivity}
Agentic LLMs are particularly sensitive to round-trip latency during continuous tool invocation, making connectivity a first-order constraint in LAUS. It is worth noting that Time-To-First-Token (TTFT) is primarily determined by prefill and inference compute rather than network transport,  rather, connectivity bounds the transport component of round-trip tool-call latency, the time required to dispatch a tool call and receive its result over the network. While 5G URLLC provides the foundational low-latency backbone for this transport layer, emerging 6G networks\cite{11433611},\cite{11373363} offer sub-millisecond latency and Integrated Sensing and Communication (ISAC) capabilities that directly address the intermittent connectivity and multi-agent synchronization challenges inherent to large-scale UAV swarms. By minimizing transport-layer delay, these advances enable LLM agents to sustain episodic memory coherence and execute real-time tool calls with fewer network-induced stalls, even as reasoning-side latency remains governed by compute rather than connectivity.

\subsection{Multimodal LLMs}

Recent work on MLLM-assisted human–swarm interaction shows that integrating MLLMs into UAV swarms significantly improves intuitive command interpretation, autonomous coordination, and operational resilience. Frameworks such as SwarmChat \cite{eumi2025swarmchat} reduce operator cognitive load by using LLM modules for context generation, intent recognition, task planning, and modality selection across natural language, voice, and teleoperation inputs, although current rule-based intent recognition requires more scalable, learning-based solutions. Additionally, GPT-4–based approaches achieve over 80\% success in formation control by jointly reasoning over visual and textual command.

\subsection{Cybersecurity}
Cybersecurity is a fundamental requirement for LAUS due to their expanded attack surface, which spans both traditional UAV network vulnerabilities and new risks introduced by LLM-assisted agentic control. Threats such as spoofing, jamming, and false data injection affect the cyber-physical layer, while prompt injection, memory poisoning, and malicious tool use target the agent’s reasoning loop, potentially leading to unsafe autonomous actions. As a result, securing LAUS requires a cross-layer defense strategy that protects communication infrastructure and ensures the integrity of perception, memory, and decision-making within the agent itself.

\begin{figure*}[t]
    \centering
    \captionsetup{justification=raggedright}
    \includegraphics[width=\linewidth]{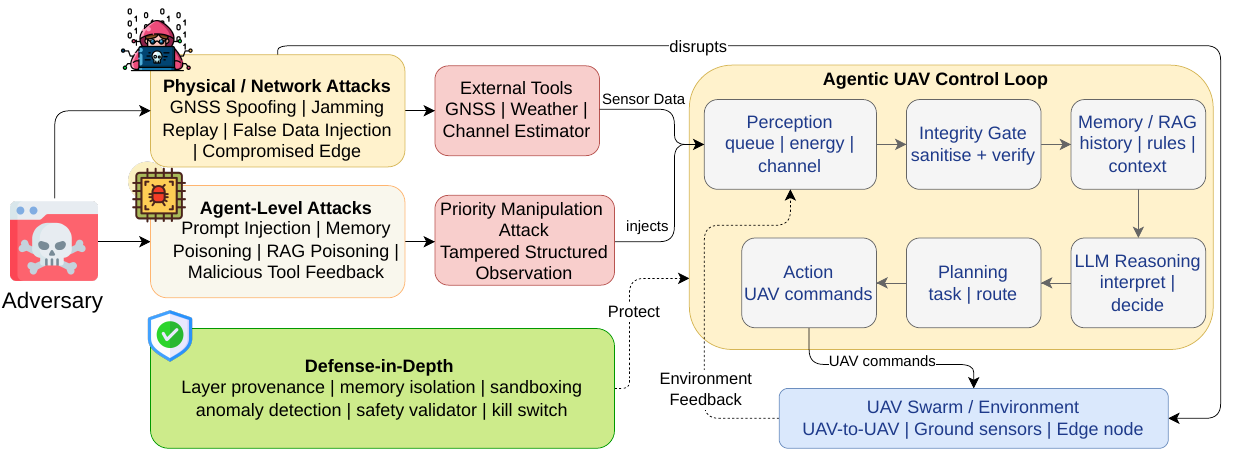}
    \caption{Multi-layered adversarial threat model and defense-in-depth
    architecture for LAUS.}
    \label{fig:digital2}
\end{figure*}

\par
While the enabling technologies discussed above provide the critical infrastructure for agentic UAV autonomy, they simultaneously expand the attack surface: each layer, perception, memory, and tool invocation, introduces new entry points for adversaries to manipulate observations, poison 
context, or subvert reasoning. Crucially, these threats do not require compromising model parameters, making them particularly insidious. The following section examines these emerging adversarial threats and outlines corresponding defense strategies to safeguard UAV swarm operations.

\section{Adversarial Threats and Defense Mechanisms} \label{sec4}

The integration of LLMs into UAV swarms brings advanced capabilities for autonomous decision-making, coordination, and adaptation, but also significantly expands the attack surface compared to conventional AI systems. Unlike traditional cyber-physical systems, LLM-assisted agents are vulnerable not only to software and communication attacks but also to manipulation of prompts, memory, retrieved knowledge, tool interactions, and perception inputs. Recent attacks on Agentic AI systems \cite{11447227} have shown that adversaries can influence autonomous behavior without compromising model parameters, underscoring the need to secure the entire decision-making pipeline, not just the LLM itself.
\par
Adversarial threats against LAUS are more complex than conventional cyberattacks because they target not only communication channels or onboard controllers, but also the agent’s reasoning, memory, tool use, and decision-making processes \cite{zhang2025agent,he2025emerged}. In an agentic UAV swarm, an LLM agent may receive multimodal observations, retrieve previous mission context, invoke external tools, communicate with other agents, and ultimately generate executable control decisions. Consequently, an adversary can manipulate the agent at various stages of the control loop, including perception, reasoning and planning, tool invocation, memory retrieval, inter-agent communication, and action execution.
\par
Fig. \ref{fig:digital2} illustrates the adversarial attack surface and defense-in-depth architecture for LAUS. It shows how both agent-level attacks, such as prompt injection, memory poisoning, RAG poisoning, and malicious tool feedback, and physical or network attacks, such as GNSS spoofing, jamming, replay, and false data injection, can affect UAV swarm operations. The figure highlights attacks at the perception–reasoning interface, where tampered structured observations may mislead the LLM reasoning and planning modules. It also presents key defensive mechanisms, including integrity verification, provenance tracking, memory isolation, sandboxing, anomaly detection, safety validation, audit logging, and kill-switch mechanisms.
\par
One important class of attacks is prompt- and context-level manipulation. In this case, adversarial instructions may be injected through user commands, retrieved documents, sensor descriptions, inter-agent messages, or tool outputs \cite{zhang2025agent,dehghan}. Such attacks can alter the agent’s interpretation of the mission, cause unsafe replanning, or bypass safety policies. For example, a malicious instruction embedded in retrieved mission data may persuade the agent to ignore no-fly-zone constraints, change task priorities, or select unsafe trajectories. In multi-agent UAV swarms, the risk increases because a compromised agent may propagate misleading context to other UAVs, resulting in coordinated but unsafe swarm behavior.
\par
A second class of attacks targets memory, RAG, and tool-using capabilities. Memory poisoning can insert malicious or misleading information into the agent’s long-term mission state, allowing the attack to persist across multiple decision cycles \cite{zhang2025agent}. RAG poisoning can corrupt the external knowledge used by the agent, leading to incorrect reasoning even when the initial user instruction is benign. Similarly, malicious tool feedback can deceive the agent by providing plausible but incorrect outputs from weather services, GNSS modules, channel estimators, or simulators \cite{dehghan}. These attacks are especially dangerous in UAV swarms because tool outputs are often directly linked to physical decisions such as velocity control, collision avoidance, resource allocation, and emergency response.
\par
A third class of threats is adversarial agentic behavior, where the agent becomes compromised, misaligned, or excessively autonomous. In this scenario, the agent may misuse tools, escalate privileges, generate unauthorized commands, or optimize for an incorrect objective \cite{he2025emerged}. For UAV swarms, this can result in unsafe route selection, inefficient sensing schedules, degraded communication quality, delayed data collection, or increased risk of physical collisions. Unlike traditional software vulnerabilities, these failures may arise from subtle interactions among probabilistic reasoning, incomplete context, tool uncertainty, and autonomous execution.
\par
Another relevant threat involves model extraction through repeated access to cloud- or edge-hosted LLM APIs. While knowledge distillation is a legitimate method for transferring knowledge from a large model to a smaller one, an adversary may collect numerous input–output pairs from the LAUS reasoning service and use them to train a substitute model that approximates its proprietary planning or decision-making capabilities. Such attacks can expose intellectual property and allow adversaries to reproduce, analyze, or exploit swarm-control behavior. Appropriate countermeasures include authenticated and least-privilege API access, adaptive rate limiting, query-pattern monitoring, abuse detection, output fingerprinting or watermarking, and restrictions on high-fidelity outputs. These controls are especially important when LAUS relies on remotely hosted LLMs accessed by multiple UAVs or edge nodes.
\par
Defending LAUS requires a defense-in-depth architecture throughout the entire agentic loop. At the input and retrieval stages, prompt sanitization, retrieval risk scoring, provenance tracking, and trusted data validation should be implemented to prevent prompt injection and RAG poisoning \cite{zhang2025agent}. At the memory level, memory isolation, write-access control, periodic memory auditing, and rollback mechanisms are necessary to prevent persistent contamination. At the tool-use level, least-privilege access, typed APIs, sandboxed execution, tool output verification, and policy-based action filters should be enforced \cite{zhang2025agent,he2025emerged}. For high-impact UAV actions, the agent’s proposed commands should be reviewed by independent safety monitors, physical constraint validators, and HITL approval mechanisms.
\par

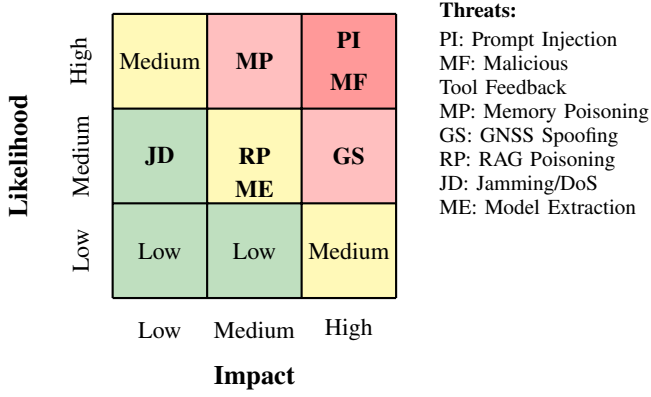
\begin{figure}[t]
\centering
\begin{tikzpicture}[
    every node/.style={font=\small},
    risk/.style={font=\small\bfseries, align=center}
]

\def\cellsize{1.25}

\fill[green!25]  
    (0,0) rectangle (\cellsize,\cellsize);
\fill[green!25]  
    (\cellsize,0) rectangle ({2*\cellsize},\cellsize);
\fill[yellow!35] 
    ({2*\cellsize},0) rectangle ({3*\cellsize},\cellsize);

\fill[green!25]  
    (0,\cellsize) rectangle (\cellsize,{2*\cellsize});
\fill[yellow!35] 
    (\cellsize,\cellsize) rectangle ({2*\cellsize},{2*\cellsize});
\fill[red!25]    
    ({2*\cellsize},\cellsize) rectangle ({3*\cellsize},{2*\cellsize});

\fill[yellow!35] 
    (0,{2*\cellsize}) rectangle (\cellsize,{3*\cellsize});
\fill[red!25]    
    (\cellsize,{2*\cellsize}) rectangle ({2*\cellsize},{3*\cellsize});
\fill[red!40]    
    ({2*\cellsize},{2*\cellsize}) rectangle ({3*\cellsize},{3*\cellsize});

\draw[thick,step=\cellsize]
    (0,0) grid ({3*\cellsize},{3*\cellsize});

\node[font=\normalsize\bfseries]
    at ({1.5*\cellsize},-1.05) {Impact};
\node[rotate=90,font=\normalsize\bfseries]
    at (-1.25,{1.5*\cellsize}) {Likelihood};

\node at ({0.5*\cellsize},-0.42) {Low};
\node at ({1.5*\cellsize},-0.42) {Medium};
\node at ({2.5*\cellsize},-0.42) {High};

\node[rotate=90] at (-0.42,{0.5*\cellsize}) {Low};
\node[rotate=90] at (-0.42,{1.5*\cellsize}) {Medium};
\node[rotate=90] at (-0.42,{2.5*\cellsize}) {High};

\node[risk] at ({2.5*\cellsize},{2.72*\cellsize}) {PI};
\node[risk] at ({2.5*\cellsize},{2.28*\cellsize}) {MF};
\node[risk] at ({1.5*\cellsize},{2.5*\cellsize}) {MP};
\node[risk] at ({2.5*\cellsize},{1.5*\cellsize}) {GS};
\node[risk] at ({1.5*\cellsize},{1.5*\cellsize}) {RP};
\node[risk] at ({0.5*\cellsize},{1.5*\cellsize}) {JD};

\node[risk] at ({1.5*\cellsize},{1.15*\cellsize}) {ME};

\node at ({0.5*\cellsize},{0.5*\cellsize}) {Low};
\node at ({1.5*\cellsize},{0.5*\cellsize}) {Low};
\node at ({2.5*\cellsize},{0.5*\cellsize}) {Medium};
\node at ({0.5*\cellsize},{2.5*\cellsize}) {Medium};

\node[
    anchor=west,
    align=left,
    font=\footnotesize
] at ({3*\cellsize+0.45},{2*\cellsize})
{
\textbf{Threats:}\\[2pt]
PI: Prompt Injection\\
MF: Malicious \\Tool Feedback\\
MP: Memory Poisoning\\
GS: GNSS Spoofing\\
RP: RAG Poisoning\\
JD: Jamming/DoS\\
ME: Model Extraction
};

\end{tikzpicture}
\caption{Qualitative risk matrix for major threats affecting LAUS.}
\label{fig:qualitative-risk-matrix}
\end{figure}

Runtime monitoring is essential. The system should continuously observe reasoning traces, tool-call sequences, inter-agent messages, command histories, and mission-level outcomes to detect abnormal behavior. In addition, tamper-evident logging, emergency kill-switch mechanisms, and fail-safe fallback controllers should be integrated to ensure that unsafe autonomous decisions do not directly result in physical harm \cite{zhang2025agent}. Securing LAUS requires shifting from model-centric security to agent-centric and mission-aware security, where every stage of perception, reasoning, memory, tool use, and action execution is treated as a potential attack surface.
\par
Fig.~\ref{fig:qualitative-risk-matrix} presents a qualitative risk matrix for the principal security threats affecting LLM-centric agentic UAV swarms, classified along two axes: exploitability (ease of mounting without physical or model-parameter access) and operational impact (degree of corruption to swarm-level decisions or physical safety). Prompt Injection (PI) and Malicious Tool Feedback (MF) occupy the highest-risk region, as they require only observation-level access yet can directly redirect UAV trajectories or override safety constraints. Memory Poisoning (MP) is rated high-impact due to its persistence across decision cycles, though slightly lower in likelihood. GNSS Spoofing (GS) and RAG Poisoning (RP) fall in the medium-to-high region given their dependence on infrastructure access, while Jamming and Denial-of-Service (JD) are assigned moderate risk, as protocol-level redundancy can partially bound their disruptive effects.
\begin{figure*}[!t]
    \centering
    \includegraphics[width=\textwidth]{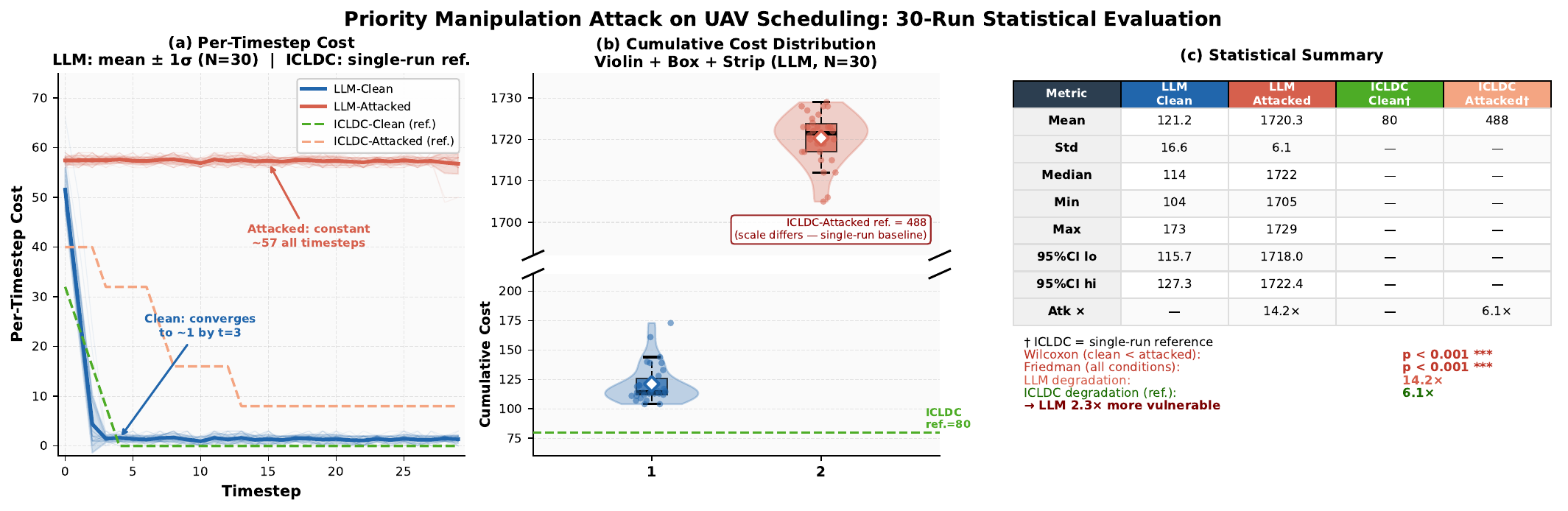}
    \caption{Performance evaluation of the LLM-assisted UAV scheduling
    agent under Priority Manipulation Attacks (PMA) over $N{=}30$
    independent runs (seeds 0--29).
    \textbf{(a)}~Per-timestep packet-loss cost: solid lines denote
    the mean; shaded bands indicate ${\pm}1\sigma$; faint traces
    show individual runs. Dashed lines correspond to the ICLDC
    scheduling scheme~\cite{11184156}.
    \textbf{(b)}~Cumulative cost distribution (violin\,+\,box\,+\,strip);
    diamond markers indicate the mean; the dashed horizontal line
    shows the ICLDC cost under clean conditions~\cite{11184156}.
    \textbf{(c)}~Statistical summary; see also
    Table~\ref{tab:statistical_summary}.
    LLM-C: clean; LLM-A: under PMA.}
    \label{fig:pma_results}
\end{figure*}

\par
The practical impact of adversarial threats on real-world swarm operations remains abstract without concrete demonstration. To bridge this gap, the following case study presents a PMA against LAUS, explicitly targeting the perception-reasoning interface.


\section{Case Study: Priority Manipulation Attacks} \label{sec5}

This case study evaluates a PMA in data-collection scheduling scenario where an LLM assigns UAVs to 20 ground sensors based on structured inputs from a perception module. The system uses perception, memory, and reasoning components to prioritize sensors according to queue length, energy, channel quality, and location to minimize packet loss. The PMA targets the perception–reasoning interface by altering sensor-level features (such as queue backlog, residual energy, and channel conditions) after perception but before LLM input, without changing the physical environment or the model itself. This gray-box attack demonstrates how subtle data manipulation at the input stage can mislead the LLM’s scheduling decisions and redirect UAV behavior, highlighting a critical vulnerability in agentic UAV systems.
\par
Unlike conventional false-data injection, PMA targets the perception–reasoning interface. The attacker has read-write access to the structured observation after it is generated by the perception module but before it is incorporated into the LLM prompt, and may manipulate selected sensor features, queue length, residual energy, and channel metrics, while leaving the physical environment, UAV trajectories, memory module, and LLM unchanged. This gray-box model reflects realistic scenarios in which an adversary compromises an edge-processing node or intercepts data within the observation pipeline.
\par
We implement a Python-based testbed that integrates GPT-4o-mini with a discrete-time kinematic UAV simulator using a probabilistic Line-of-Sight channel model. The proposed LLM agent is evaluated over $N{=}30$ independent runs with seeds 0–29. The ICLDC scheduling scheme~\cite{11184156} serves as a baseline for comparison under the same environmental configuration. At each timestep, the perception module interfaces with the LLM via structured JSON prompts that encode real-time sensor state. Under PMA, attacker-manipulated features are injected before prompt construction. Table~\ref{tab:pma-parameters} summarizes the evaluation parameters.  Each ground sensor node is initialized with 50 J of energy at simulation start; The attack Intensity is applied to the three input features: path loss, energy level, and queue length.

\begin{table}[!t]
\centering
\caption{PMA Evaluation Parameters}
\label{tab:pma-parameters}
\footnotesize
\renewcommand{\arraystretch}{1.15}
\begin{tabular}{lc}
\toprule
\textbf{Parameter} & \textbf{Value} \\
\midrule
LLM Model              & GPT-4o-mini \\
LLM temperature        & 0.2 \\
Number of UAVs         & 3 \\
Number of Sensors      & 20 \\
Simulation Area        & $100 \times 100~\text{m}^2$ \\
UAV Altitude           & $100~\text{m}$ \\
Initial Energy         & $50~\text{J}$ \\
Simulation Horizon     & 30 timesteps \\
\midrule
Independent Runs ($N$) & 30 \\
Random Seeds           & 0--29 \\
Statistical Test       & Wilcoxon signed-rank \\
Attack Target Sensors  & \{2,\,7,\,15\} \\
Attack Intensity       & $3\times$ \\
ICLDC Baseline         & \cite{11184156} \\
\bottomrule
\end{tabular}
\end{table}

Fig.~\ref{fig:pma_results} and Tables~\ref{tab:statistical_summary} present the results.
The LLM agent achieves a mean cumulative packet-loss cost of 121.2 (95\,\% CI: [115.7,\,127.3], $\sigma{=}16.6$) under clean conditions,
rising to 1720.3 under PMA --- a $14.2{\times}$ degradation ratio. A Wilcoxon signed-rank test confirms this difference is statistically significant (p<0.001), with a 14.2× degradation ratio in cumulative packet-loss cost (95\% CI: [115.7, 127.3] clean vs. [1718.0, 1722.4] attacked). The ICLDC baseline achieves a cumulative cost of 80 under clean
conditions and 488 under attack, corresponding to a $6.1{\times}$ degradation. The PMA was effective in all 30 LLM runs: at every timestep, all three UAVs were redirected exclusively toward sensors
\{2,\,7,\,15\}, confirming that observation-level manipulation alone suffices to deterministically  steer LLM scheduling decisions without any access to model parameters.

\

\begin{table}[!t]
\caption{Cumulative Packet-Loss Cost: LLM Agent ($N{=}30$)
         vs.\ ICLDC Baseline~\cite{11184156}.}
\label{tab:statistical_summary}
\centering
\scriptsize
\setlength{\tabcolsep}{3pt}
\renewcommand{\arraystretch}{1.1}
\begin{tabular}{lcccc}
\toprule
\textbf{Metric}
  & \textbf{LLM-C}
  & \textbf{LLM-A}
  & \textbf{ICLDC-C}
  & \textbf{ICLDC-A} \\
\midrule
Mean             & 121.2 & 1720.3 & 80   & 488  \\
$\sigma$         & 16.6  & 6.1    & {--} & {--} \\
95\,\%\,CI\,(lo) & 115.7 & 1718.0 & {--} & {--} \\
95\,\%\,CI\,(hi) & 127.3 & 1722.4 & {--} & {--} \\
Atk.\ ratio      & {--}  & \textbf{14.2$\times$}
                          & {--}   & \textbf{6.1$\times$} \\
\midrule
Wilcoxon (C\,$<$\,A)
  & \multicolumn{4}{c}{$p < 0.001$\quad(***)\quad[LLM, $N{=}30$]} \\

\bottomrule
\end{tabular}
\end{table}

These findings confirm that manipulating only the structured observation is sufficient to influence LLM-generated control actions and degrade system performance, highlighting the vulnerability of the
perception--reasoning interface in LAUS.

The 30-run evaluation establishes PMA as a statistically robust and practically accessible threat. The $14.2{\times}$ degradation in
cumulative packet-loss cost (Fig.~\ref{fig:pma_results}a,
Table~\ref{tab:statistical_summary}), replicated without exception across all independent runs (p<0.001, 14.2× degradation ratio),
confirms that observation-level access alone suffices to systematically redirect LLM scheduling decisions without any knowledge of model weights or architecture. This positions the perception--reasoning
interface as a high-leverage, low-barrier attack surface that bypasses conventional model-centric defenses entirely.
═══════════════════════════════════════════════════════════════════════
\section{Discussion and Open Problems} \label{sec6}
Several open problems remain. Hallucination-resistant reasoning and standardized output verification for agentic UAV controllers are still lacking. Onboard deployment under strict SWaP constraints remains unresolved, as current compression techniques reduce the reasoning depth required for high-complexity missions. Standardized security benchmarks for perception-reasoning attacks do not yet exist; the PMA presented here illustrates one vector, but a systematic taxonomy is needed. Finally, multi-agent isolation protocols that contain a compromised agent's influence without suspending swarm operations remain an open engineering challenge.
\par
Beyond these open problems, critical operational challenges persist, hindering the full deployment of LAUS in UAV swarms. These include inherent difficulties in complex multi-agent coordination, where inter-agent dependencies and partial observability complicate robust collective behavior. Instability in task planning remains prevalent, with frequent replanning cycles and brittle execution under dynamic environmental conditions. In addition, reliance on external knowledge bases introduces significant bottlenecks: low retrieval efficiency leads to increased latency and context misalignment, degrading real-time decision-making performance, particularly in bandwidth-constrained or contested communication environments. Furthermore, external tool integration and invocation remain insufficiently flexible, limiting seamless interaction between the LLM-assisted decision layer and heterogeneous onboard or edge-resident control, sensing, and actuation modules \cite{11370176}.
\par
A key operational challenge is the constraint imposed by the UAV platform itself: a primary research priority for LAUS is enabling efficient deployment of LLM-assisted agents on resource-constrained UAV platforms. Current UAVs have limited onboard CPU, GPU, memory, and battery capacity, making direct execution of LLMs impractical; moreover, the high computational and energy demands of these models can significantly reduce flight endurance. Future work should focus on developing lightweight agentic architectures using model compression, pruning, quantization, knowledge distillation, and Small Language Models (SLMs) tailored to UAV applications.
\par
Closely related to this deployment constraint is the challenge of real-time edge inference, which remains significant for LAUS. Unlike conventional deep neural networks, LLM inference is stateful, requires continuous management of growing KV caches, and involves distinct prefill (compute-bound) and decoding (memory-bound) phases. Accordingly, research should focus on efficient KV-cache management, phase-aware scheduling, and adaptive resource allocation to reduce latency and energy consumption while preserving inference quality.

Taken together, these architectural and deployment challenges underscore the need for rigorous evaluation methodologies. Research should therefore focus on developing systematic reliability evaluation frameworks that assess adversarial robustness, hallucination, bias, and safety in a unified manner using standardized benchmarks. Another direction is the design of domain-specific reliability benchmarks for safety-critical applications such as UAV communications, incorporating realistic conditions such as noise, incomplete inputs, and adversarial scenarios.

\section{Conclusion} \label{sec7}
This paper proposed LAUS and reviews the key technologies required to realize this vision. We further examined emerging adversarial threats targeting agentic UAV systems, and discussed defense-in-depth strategies. Through a case study on a PMA, we demonstrate that manipulating structured observations at the perception–reasoning interface can significantly influence LLM-generated decisions and degrade swarm performance. Finally, the success of LAUS will depend not only on improving the intelligence of individual agents but also on developing secure, transparent, and accountable agentic ecosystems that can be trusted to operate in safety-critical environments. 

\bibliographystyle{IEEEtran}
\bibliography{references}
\end{document}